\documentclass{article}

    \usepackage[nonatbib, preprint]{neurips_2022}

\usepackage[utf8]{inputenc} % allow utf-8 input
\usepackage[T1]{fontenc}    % use 8-bit T1 fonts
\usepackage{url}            % simple URL typesetting
\usepackage{booktabs}       % professional-quality tables
\usepackage{amsfonts}       % blackboard math symbols
\usepackage{nicefrac}       % compact symbols for 1/2, etc.
\usepackage{microtype}      % microtypography
\usepackage{xcolor}         % colors

\usepackage{subfig}

\usepackage[utf8]{inputenc}
\usepackage{mathtools}
\usepackage{array}
\usepackage{booktabs, multirow} % for borders and merged ranges
\usepackage{soul}% for underlines
\usepackage{graphics}
\usepackage{graphicx}
\usepackage{multicol}
\usepackage{caption}
\newcommand{\change}[1]{\textcolor{black}{#1}}

\title{Rich Feature Distillation with Feature Affinity Module for Efficient Image Dehazing}

\author{%
  Sai Mitheran J., Anushri Suresh, Nisha J. S., Varun P. Gopi\thanks{Corresponding Author} \\
  Department of Electronics and Communication
Engineering\\
  National Institute of Technology Tiruchirappalli
 \\
}

\begin{document}

\maketitle

\begin{abstract}
Single-image haze removal is a long-standing hurdle for computer vision applications. Several works have been focused on transferring advances from image classification, detection, and segmentation to the niche of image dehazing, primarily focusing on contrastive learning and knowledge distillation. However, these approaches prove computationally expensive, raising concern regarding their applicability to on-the-edge use-cases. This work introduces a simple, lightweight, and efficient framework for single-image haze removal, exploiting rich “dark-knowledge" information from a lightweight pre-trained super-resolution model via the notion of heterogeneous knowledge distillation.  We designed a feature affinity module to maximize the flow of rich feature semantics from the super-resolution teacher to the student dehazing network. In order to evaluate the efficacy of our proposed framework, its performance as a plug-and-play setup to a baseline model is examined. Our experiments are carried out on the RESIDE-Standard dataset to demonstrate the robustness of our framework to the synthetic and real-world domains. \change{The extensive qualitative and quantitative results provided establish the effectiveness of the framework, achieving gains of upto 15\% (PSNR) while reducing the model size by $\sim$20 times.}

\end{abstract}

\section{Background}
The presence of atmospheric particles, fog, mist, smoke, etc., causes absorption and scattering of light, which in turn results in what is known as the haze. This haze is responsible for poor visibility conditions, including those captured on cameras. Images captured in hazy and bad weather conditions are of poor quality, with degraded color and details inhibiting clear visibility of the scene. The necessity for removing haze and fog from these images, i.e., dehazing, is manifold. Not only would we benefit from more transparent images for day-to-day activities, but several computer vision tasks, including object detection, recognition, tracking, and segmentation, rely on clean, high-quality images as inputs. Thus the problem of dehazing becomes a pre-processing step necessary in carrying out computer vision experiments.  

The advent of deep learning in computer vision tasks has not gone unnoticed. The breakthroughs of convolutional neural networks \cite{3a6e6cc3ac6e4326b35a6315d7cbddf7} in their diverse use, and groundbreaking accuracy have been exploited too many problems. Several deep learning approaches have been used to tackle haze and fog removal in the past \change{\cite{DBLP:conf/aaai/QinWBXJ20, Ren2016SingleID, Gui2021ACS,ICCV17a, liuICCV2019GridDehazeNet, DBLP:journals/tip/ZhangT20,Cai2016DehazeNetAE, 8953692, engin2018cycle}}. However, the need for heavy resources, compute, time and storage makes this approach problematic. One method in supervised learning is to use the Atmospheric Scattering Model (ASM) \cite{1069864, 1201821} in collaboration with deep learning to estimate the parameters (i) $t(x)$ - Trans and (ii) $A$ - atmospheric light value in the following equation:

\begin{equation}
I(x)=J(x) t(x)+A(1-t(x))
\end{equation}

Here, $x$ represents the position of pixels. $I(x)$ denotes the hazy image, and $J(x)$ represents the clear image that the dehazing method expects to obtain. Strategies described previously have used parameter estimation, such as multi-scale \cite{10.1007/978-3-319-46475-6_10}, pyramid network \cite{8578435}, monocular estimation \cite{9157136}. 
However, the drawback of this method is that the estimated values of both t(x) and A are merely estimations/approximations. This incurs errors in the final image obtained, thus failing to produce the desired result. 

Previous supervised learning work has also explored dehazing methods independent of the ASM, i.e., without estimating t(x) and A. These methods attempt to directly find a mapping between the hazy and clear images using several models such as \cite{Ren2018GatedFN, DBLP:conf/aaai/QinWBXJ20, 8953692, 9156318, liuICCV2019GridDehazeNet, DBLP:journals/corr/abs-2005-04668}. The drawback of this direct mapping is the loss of information in the process. An issue posed by supervised learning is the need for pairs of hazy and clear images. This may not be obtainable in all cases, especially in the case of outdoor images, where a clear and hazy image with the presence of plants, trees, water, and the sky is difficult to obtain. Hence unsupervised dehazing methods find themselves useful in this situation. Previous methods have scratched the surface of semi-supervised and unsupervised learning. The use of GANs \cite{NIPS2014_5ca3e9b1, engin2018cycle, Zhu2017UnpairedIT, Li_2018_CVPR}, image to image translation \cite{8953692, Isola2017ImagetoImageTW} have been explored in the past. However, the inherent difficulty of this method makes it relatively unsuccessful. There is also significantly less work in this domain.

% To add a paragraph here, introducing our work

% open problems (other than unsup)
% what are we trying to tackle
% this work focuses on on the edge dehazing
% applications of on the edge
% highlight our work and how it solves problems

\change{Besides the instability of training GANs, other open areas exist for research in single image haze removal. Self-supervised learning (SSL) for image dehazing~\cite{Wu2021ContrastiveLF} is yet to be explored in-depth primarily due to large pre-training compute requirements. Real-world data domain adaptation still needs improvement. Furthermore, there does not exist much work on computationally cheap models~\cite{article, DBLP:journals/tip/ZhangT20} with faster inference time and reduced size. With rapidly increasing on-the-edge applications of haze removal in remote sensing, surveillance, and vehicular systems, the compute overhead of heavy models~\cite{Zhang_2018_CVPR, https://doi.org/10.48550/arxiv.2004.13388, engin2018cycle, 9578433} need to be minimized while maximizing their performance. This is our primary motivation in introducing this work, leveraging rich super-resolution features to adaptively guide our lightweight dehazing network through a novel feature affinity module.}

\change{The main idea of this paper is to design an efficient, lightweight pipeline for single image haze removal by exploiting a rich pre-trained teacher super-resolution teacher network to distill multi-scale information to the student dehazing network. Using our feature affinity module, the fine-grained multi-scale features are merged into the student's feature distribution. Our key insight is to transfer the potential of knowledge distillation across two heterogeneous tasks while exploiting fine details in high-resolution images after super-resolution. The increased pixel density in the clear, high-resolution counterpart of the hazy images provides rich semantics to study the uneven haze distribution, which our lightweight student effectively models. Our experiments on the RESIDE-Standard test set show that at inference time, our student network decoupled from the framework can successfully remove haze from images at reduced computational loads and improved performance. We achieve gains of up to 15\% (PSNR) while reducing the model size by $\sim$20 times, thus ensuring computational efficiency, cheap inference cost, and reasonable performance to accelerate research in the direction of improved on-the-edge methods for dehazing.} 

The contributions of our work can be summarized as follows:

\begin{itemize}
\item Proposed an efficient, lightweight framework for the task of single image haze removal from indoor and outdoor scenes.
    \item Designed a lightweight student network for single image haze removal\change{, incorporating cascaded blocks of residual connections, pixel-level and channel-level attention to fuse semantics across all dimensions and learn a rich intermediate feature representation.}
    \item \change{Introduced a novel} feature affinity module to utilize rich super-resolution features from a pre-trained network. This distills multi-scale information to the student dehazing network to provide "dark knowledge supervision\footnote{\change{Dark Knowledge Supervision refers to the hidden (dark) information of clear images as learned by the super-resolution teacher network in the form of intermediate multi-scale features. Distilling this knowledge to the lightweight student dehazer provides implicit guidance, enabling the smaller model to better study the haze distribution on images.}}," guiding the feature representation learned by the dehazing network. \change{The results obtained on the RESIDE-Standard dataset solely adding our feature affinity module and rich teacher features demonstrate the improved performance at minimal computational overhead.}

\end{itemize}

\section{Related works}
\subsection{Single Image Haze Removal}

Several methods exist in the paradigm of Supervised Learning~\cite{Cai2016DehazeNetAE, liuICCV2019GridDehazeNet, Zhang_2018_CVPR, ICCV17a, DBLP:journals/tip/ZhangT20, DBLP:conf/aaai/QinWBXJ20, DBLP:conf/cvpr/HongXLQ20, 9156566, 8953692, Li_2018_CVPR} and Unsupervised Learning~\cite{engin2018cycle, 10.1145/3394171.3413876, 8803316, 10.1109/TIP.2020.3007844, Yang_Xu_Luo_2018, 8897130} to remove haze from images. Early works in the single image haze removal regime focused on the ASM, commonly referred to as the physical scattering model~\cite{1069864, 1201821}. The motivation behind the ASM is to estimate the transmission map $t(x)$ and the global atmospheric light $A$, from which the dehazed image can be evaluated analytically. As the use of deep neural networks for function approximation~\cite{Liang2017WhyDN} became widespread, ASM-based supervised methods used neural networks to either approximate $t(x)$ with empirical hand-crafted priors for $A$~\cite{Cai2016DehazeNetAE, Ren2016SingleID} or exploited joint-estimation of $t(x)$ and $A$~\cite{Zhang_2018_CVPR} with multiple neural networks. Since these function approximation methods are associated with empirical assumptions and prone to element-wise errors for the components of $t(x)$ and $A$, the errors compound on the joint estimation result in a higher degree of inaccuracy. As a result, satisfactory haze-free outputs are obtained, introducing additional artifacts. \change{The Structural Similarity Index Measurement (SSIM) obtained for the end-to-end deep learning-based dehazing method was 0.9776. However, the dense network maximizes the information flow from different levels. It reduces the mutual structural information between the estimated transmission map.}

Moving away from estimating $t(x)$ and $A$, \cite{ICCV17a, DBLP:journals/tip/ZhangT20} introduced an additional parameter to be estimated, accounting for the information content of the two ASM parameters. As a result, the dehazed image can be obtained with accumulated uncertainties only from a single parameter, not requiring the expensive transmission map and atmospheric light data. \change{In \cite{ICCV17a} the experiments were conducted on two datasets, namely TestSet A and B.  Average PSNR and SSIM results on TestSet A were 19.6954 and 0.8478, respectively. Average PSNR and SSIM results on TestSet A were 21.5412 and 0.9272, respectively. The proposed CNN model was successfully used to remove haze.
However, atmospheric light can not be regarded as a global constant to be learned alongside medium transmission in a unified network.} Since most of these methods were not end-to-end~\cite{ICCV17a} and as learning-based image translation emerged, newly proposed pipelines~\cite{DBLP:conf/aaai/QinWBXJ20, DBLP:conf/cvpr/HongXLQ20} adopted strategies to map directly between the hazy and clear image distributions.

GFN~\cite{Ren2018GatedFN} modeled haze removal as the weighted combination of the images’ white-balanced, contrast-enhanced, and gamma-corrected derivatives. However, this method was associated with a hand-crafted pre-processing pipeline, which could be eliminated. \change{The main limitation of DFN was that it could not handle corrupted images with extensive fog.} EPDN~\cite{8953692} applied for advances from the Pix2Pix image-to-image translation~\cite{Isola2017ImagetoImageTW} method using Generative Adversarial Networks (GANs)~\cite{NIPS2014_5ca3e9b1} and an enhancer. \change{EPDN produced PSNR and SSIM on outdoor datasets of 22.57 and 0.8630. For heavily hazed scenes, the proposed method was not very robust. Object edges that are heavily hazed are unable to be recovered naturally. The limitation could be overcome by incorporating more enhancing blocks into the EPDN network.} This produced visually detailed and perceptually pleasing images using a joint training scheme. However, the convergence of four loss functions and the unstable training of GANs opened the scope for improvement. GANs also led the way into unsupervised single image haze removal, as initiated by the Cycle-Dehaze network~\cite{engin2018cycle}. \change{This method for single image dehazing problems does not require training with pairs of hazy and corresponding ground truth images. The Cycle-Dehaze network produced a PSNR value of 19.62 and SSIM of 0.67.} The proposed cyclic perceptual consistency loss, regularizing the GAN-based adversarial objective used to optimize CycleGAN~\cite{Zhu2017UnpairedIT} provided the means of mapping between the clear and hazy domains without label supervision. \change{CycleGAN faces a significant challenge in handling more complex and radical transformations, particularly geometric alterations.} Building on generative modeling, DDN~\cite{Yang_Xu_Luo_2018} disentangled images to design a generator for each clear image, transmission map $t(x)$, and atmospheric light. \change{DDN produced PSNR and SSIM on the Middlebury dataset of 14.9539 and 0.7741.}  Deep-DCP~\cite{8897130} introduced an energy-function-based optimization method, learning the underlying transformation using just hazy images, minimizing the dark channel prior (DCP). Several works focused solely on image reconstruction objectives with direct supervision and no feature-level regularization. For instance, GridDehazeNet~\cite{liuICCV2019GridDehazeNet} and FFA-Net~\cite{DBLP:conf/aaai/QinWBXJ20} introduced the attention mechanism for feature sharing at different scales. 

GridDehazeNet~\cite{liuICCV2019GridDehazeNet} proposed a modular pipeline with learnable pre-processing, an attention-based backbone, and a post-processing module.
\change{GridDehazeNet produced PSNR and SSIM on the SUN RGB-D dataset of 28.67 and 0.9599. However, the dehazing method proposed here does not rely on the atmosphere scattering model.} FFA-Net~\cite{DBLP:conf/aaai/QinWBXJ20} combined spatial and channel information using attention, and adaptively learning features at different levels to capture thick-haze regions. These methods are associated with heavy computational requirements and inference costs, hindering edge applications in autonomous vehicles. Utilizing additional task information for haze removal was recently explored in~\cite{DBLP:conf/cvpr/HongXLQ20}. An off-the-shelf auto-encoder was leveraged as the teacher, performing the task of image reconstruction. The intermediate teacher features were used as a guiding signal to obtain a similar feature distribution for the dehazing student network using a process-oriented learning mechanism. 

Focusing on the domain generalization abilities of single image haze removal methods to work with real-world data, Shao et al.~\cite{9156566} proposed a domain adaptation technique to train using real hazy images. Depth information is embedded into an image translation framework between the real and synthetic domains, along with a domain-specific dehazing network, which is regularized using a consistency loss. The extension of this method into the unsupervised paradigm remains open for exploration. As most of these methods focus on improving performance by network architecture complexity and design, the self-supervised representation learning method of Contrastive Learning~\cite{pmlr-v119-chen20j} was incorporated in~\cite{Wu2021ContrastiveLF}. \change{However, this method ensures that the restored image is drawn closer to the clear image and pushed away from the hazy image while taking longer to train. It produced a PSNR and SSIM of 27.76 and 0.9284 on synthetic datasets.} Contrastive regularization with a compact auto-encoder-based network design enabled adequate information flow, using hazy images as negative samples. Their clear counterparts were used as positive reinforcements, pushing the reconstructed haze-free image towards the positive samples. In addition to these methods, this work explores the effective usage of features in a rapidly growing receptive field with feature space guidance.

\change{In \cite{zhu2020novel} an image fusion-based algorithm to improve image dehazing accuracy and reliability was proposed. To guide the fusion process, pixel-wise weight maps were constructed from gamma-corrected underexposed images by analyzing global and local exposedness. This process reduces the spatial dependence of the fused image's luminance and balances its colour saturation. The detection confidence of the obstacles after dehazing was 0.2533, which was very high compared to existing methods. The work proposed in \cite{zheng2020image} was based on the new technique of adaptive structure decomposition integrated multi-exposure image fusion (PADMEF). First, a group of underexposed images taken was extracted from a single blurred image using gamma correction and spatial linear saturation adjustment. Then, a multi-exposure image fusion scheme-based adaptive structure decomposition was used for each image patch to fuse different exposure-level images into a haze-free image. Remote sensing images have been used extensively in the military, national defence, disaster response, and environmental monitoring, among other applications. SSIM AND FADE scores for the proposed PADMEF were 0.7816 and 0.3580, respectively.}

\change{Haze removal from remote sensing images based on a dual self-attention network was proposed in \cite{zhu2021remote}. Dual self-attention defogging networks utilized residual octave convolutions, by which a source image was decomposed into high and low-frequency components. The SSIM and PSNR values obtained were 0.8661 and 34.9675. The total time taken for the dual self-attention network to remove haze from images was 0.0315s. \cite{zhu2021atmospheric} proposed a novel atmospheric light estimation-based dehazing algorithm. The colours of the images dehazed by the proposed method were consistent with human vision in various scenes. One hundred remote sensing images from hazy scenes were created for testing. The average processing time of the suggested solution was relatively high because of the network's increased time complexity. An adaptively spatial parallel convolution module was then proposed in \cite{qi2022small} to address the lack of spatial information. The split-fusion sub-module proposed in this work reduced the adaptively spatial parallel convolution module's time complexity. The suggested SODNet only improves the performance of small object detection.}

\subsection{Knowledge Distillation}
Since a principal application of single image haze removal is for on-the-edge use-cases for autonomous driving and control, the proposed method must ensure computational efficiency, cheap inference cost, and reasonable performance. Several methods in existing literature focus on model compression, such as pruning~\cite{Liu2019RethinkingTV}, knowledge distillation~\cite{Hinton2015DistillingTK}, etc. Knowledge distillation employs a larger expert teacher model, usually pre-trained for the same task, to provide control signals by distilling information to a compressed, compact student network. As the smaller student model~\cite{Hinton2015DistillingTK} trains, it is forced by the objective function to memorize the trained teacher's feature distribution implicitly. 

Knowledge distillation has come up over the years in semantic segmentation~\cite{8954081, Xie2018ImprovingFS, He2019KnowledgeAF}, image classification~\cite{Hinton2015DistillingTK, Xie2020SelfTrainingWN, Yuan2019RevisitKD}, and object detection~\cite{8100259, zhang2021improve, Wang2019DistillingOD}. For instance, \cite{8954081} introduces pair-wise and holistic distillation to ensure pixel-level, feature-level, and high-order information exchange for dense prediction. \cite{wang2020ifvd} focuses on intra-class feature variation mimicking, enforcing a similar feature distribution between class-specific pixels and a class-specific prototype in an embedding space to segment inputs accurately. \cite{Yuan2019RevisitKD} uses distillation as a regularization method, drawing parallels with label smoothing~\cite{Mller2019WhenDL}. The student model is hypothesized to learn from itself or from a virtual teacher distribution for image classification. Similarly, approaches for object detection focused on consistent feature-map distributions~\cite{8100259} between student and teacher, attention-based student model guidance~\cite{zhang2021improve}, to weigh important foreground objects higher, and down-weigh noisy background information with minor semantics.

There exists a wide range of possibilities to leverage heterogeneous task distillation to transfer knowledge from a specialized teacher on an easier task to a student target model for a relatively more complex task. KDDN~\cite{DBLP:conf/cvpr/HongXLQ20} trains a teacher model to reconstruct images and adopts the process-oriented learning method to distill knowledge through features for the student model. This imitation learning setup enables the student model to effectively remove haze to enforce more attention to the dense, hazy regions of the spatial map. Extending the capabilities of the expert teacher model, an efficient pipeline to learn richer features at the teacher is proposed, with a rapidly growing receptive field, to obtain the super-resolution of the clear image rather than simply reconstructing the input. We hypothesize that the fine-grained features at different scales, as extracted from the super-resolution network, contain richer information in the form of “dark knowledge" supervision that the student can exploit. Furthermore, we also design an effective feature affinity module at each scale to ensure that spatial and channel-aware features are distilled to make the best use of the higher resolution maps.

\section{Proposed Method}

\begin{figure}[htbp]
\begin{center}
\includegraphics[scale=0.8]{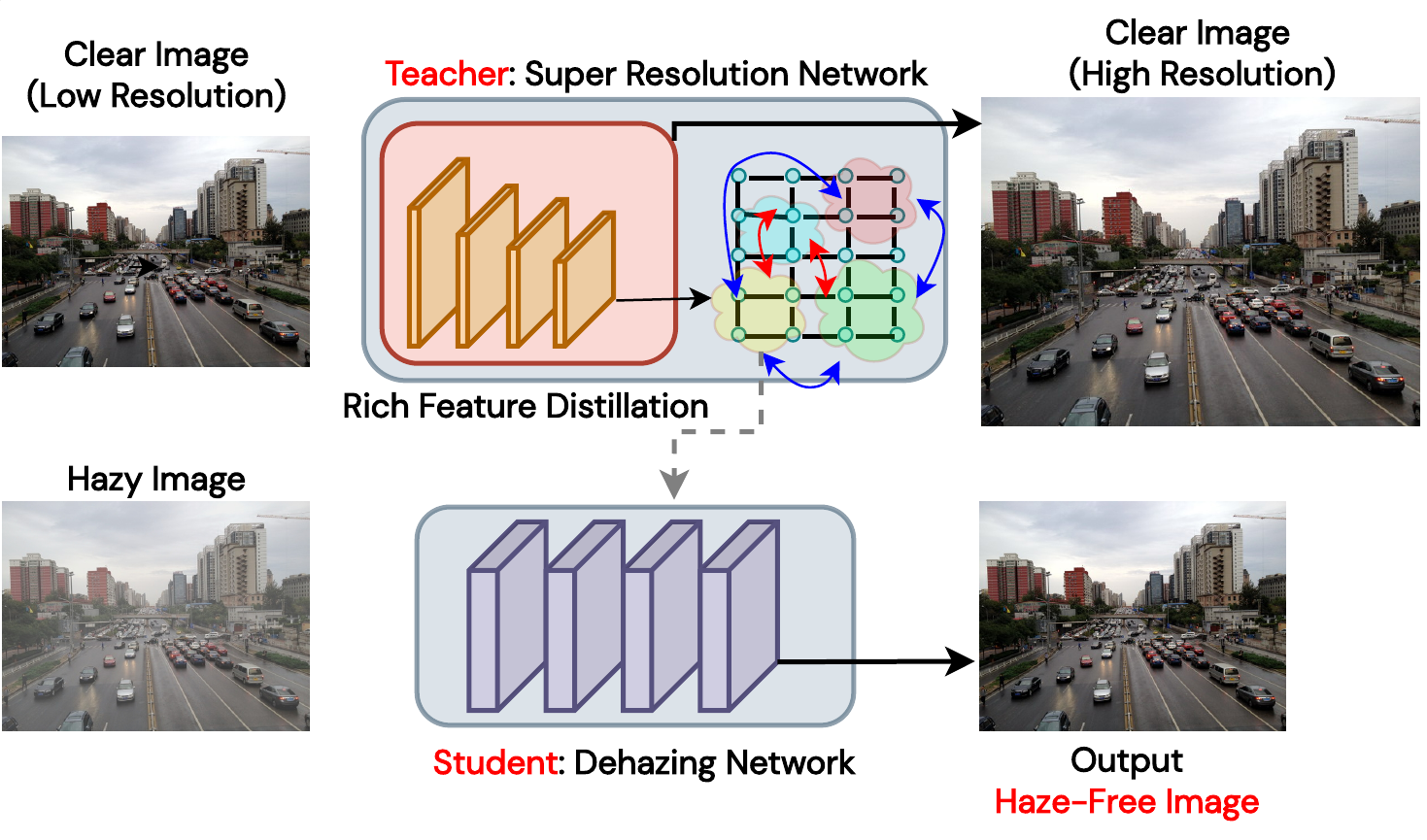}
\caption{ The proposed knowledge distillation framework for single image haze removal} 
\label{fig:main_method}
\end{center}
\end{figure}

This work proposes to exploit a rich \textit{pre-trained super-resolution network} as a teacher model, as shown in Figure~\ref{fig:main_method}. We hypothesize employing \textit{knowledge distillation}, where a larger expert teacher model, usually pre-trained for the same task, provides control signals by distilling information to a compressed, compact student network. Building on the claims of KDDN~\cite{DBLP:conf/cvpr/HongXLQ20}, this work further extends the paradigm of heterogeneous knowledge distillation, motivated by the feature affinity method proposed by Wang et al.~\cite{9157434}. Rather than using the reconstruction objective as KDDN~\cite{DBLP:conf/cvpr/HongXLQ20} for the teacher, which aids in rich feature-space information flow to the student, we aim to signify the importance of having richer teacher features for the guiding signal. For this purpose, a lightweight super-resolution network with depth-wise separable convolution is developed, which has proven to be significantly cheaper to compute than standard convolution~\cite{Howard2017MobileNetsEC}. 

As shown in Figure~\ref{fig:main_method}, the lightweight pre-trained teacher network with rich super-resolution features distills knowledge to light \textit{student dehazing network}. We hypothesize that the fine-grained features at different scales, as extracted from the super-resolution network, contain richer information in the form of “dark knowledge" supervision that the student can exploit. Finally, an intelligent \textit{feature affinity module} is designed at each scale to ensure that spatial and channel-aware features are distilled to make the best use of the higher resolution maps.

\subsection{Image Super-Resolution: Teacher Network}

Single image super-resolution refers to the process of enhancing the resolution of an image from Low-Resolution (LR) to High-Resolution (HR). The LR image is a down-sampled version of the HR image, with various other factors of degradation that affect practical use-cases, such as blur and noise. Various hand-crafted models were proposed \cite{glasner2009super, farsiu2004fast} to enable the restoration of HR images from their LR counterparts. As deep neural networks exhibited great success across several image-level tasks, super-resolution methods based on convolutional neural networks (CNN) achieved state-of-the-art performance by surpassing these hand-crafted pipelines \cite{Dong2016ImageSU}.
\vspace{0.2em}

SRCNN \cite{Dong2016ImageSU} was one of the first methods to leverage deep learning for super-resolution. It consists of three sub-modules: Patch Extraction, Linear Mapping, and Reconstruction. It involves a pre-processing stage of bicubic interpolation to the HR space, after which feature extraction occurs. Although SRCNN achieved impressive results, the pre-upsampling super-resolution process that required feature extraction in the HR space proved to be unnecessarily computationally expensive. Moving into the paradigm of post-upsampling super-resolution, FSRCNN \cite{10.1007/978-3-319-46475-6_25} intelligently extracts features in the LR space, hence reducing memory and compute requirements. Furthermore, FSRCNN exploits learned deconvolutional filters instead of the bicubic interpolation from SRCNN and simplifies the architecture's reliance on computing using cheap $1$x$1$ convolutions.

\begin{figure}[htbp]
\begin{center}
\includegraphics[width=13 cm,height=7.1 cm]{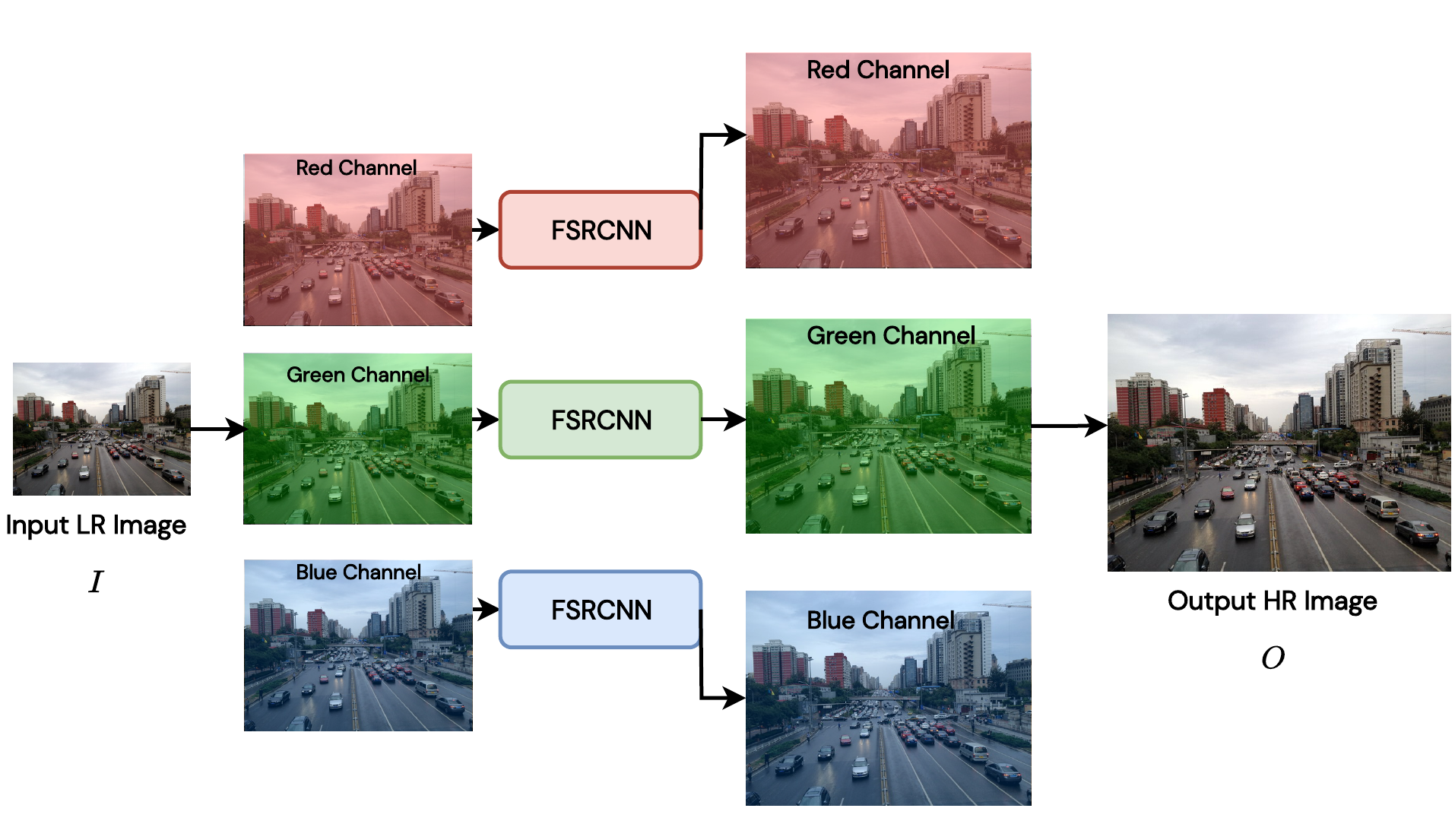}
\caption{ Proposed channel-wise multi-branch teacher network for single-image super-resolution. Fast Super-Resolution Convolutional Neural Network (FSRCNN) is adopted from~\cite{10.1007/978-3-319-46475-6_25}.} 
\label{fig:teacher}
\end{center}
\end{figure}

Extending the FSRCNN module \cite{10.1007/978-3-319-46475-6_25}, a multi-branch network is designed to carry out channel-level super-resolution of the clear input images from the dataset, as shown in Figure~\ref{fig:teacher}. \change{Given an input image $I$ with 3 channels: red, green, and blue ($R, G, B$), each channel $I^{c}$ passes through the channel-specific FSRCNN module, ${FSRCNN}_{c}$\begin{equation}
{O^{c} = FSRCNN_{c}({I}^{c})} \ 
\end{equation}}

\change{The outputs $O^{c}$ from each channel-specific FSRCNN module are then concatenated to form thee overall output $O$, as given below.}
\change{\begin{equation}
O = Concat(O^{c} )_{ c  \in (R,G,B)}
\end{equation}}
The cascaded processes of feature extraction, shrinking, mapping, expanding, and deconvolution are applied to each channel: red, green, and blue. Rather than optimization using the Mean Squared Error (MSE) objective at the image level, it is applied at the channel level to train the multi-branch network end-to-end with gradients back-propagated across each branch. The proposed alteration in the network architecture provides a considerable performance gain, as measured in Peak Signal-to-Noise Ratio (PSNR) while trading off with computing requirements. Since the architecture is replicated across N branches ($N=3$), the number of parameters and memory requirements are increased by N times. 

While attaining a performance boost for single-image super-resolution, the compute requirement for our multi-branch framework is also reduced using \textit{Dpth-wise Separable Convolutions} (DSC) \cite{8099678}. Depth-wise separable convolutions consist of two primary components: point-wise convolutions and depth-wise convolutions. These operations help significantly reduce the computational complexity while running the model on input images, as shown in Table. We experimentally verified that depth-wise separable convolutions significantly reduce the computational cost, with only a minimal reduction in PSNR. Adding this single component into the network boosts time complexity, hence making the model well suited for real-time processing and downstream applications.

\subsection{Single Image Haze Removal: Student Network}

Single image dehazing (haze removal) refers to the process of removal of haze from the input image of a hazy scene. The hazy input image is associated with low visibility, weather, dust, and other attributes that degrade the image's original quality for further processing required in downstream tasks. In a standard haze removal framework, the network's output is associated with improved visibility and void of haze, making it clear for computer vision applications. 

\begin{figure}[htbp]
		\centering
		\includegraphics[width=14 cm,height=4.75 cm] {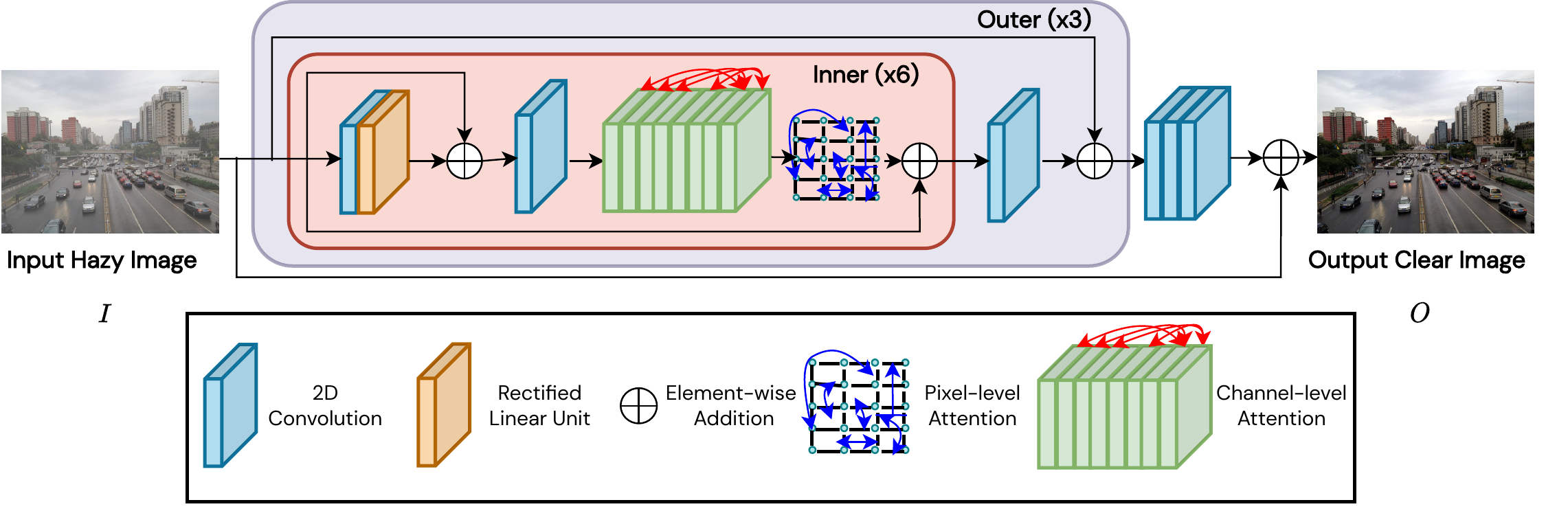}
		\caption{\change{Student network for single-image haze removal}}
		\label{fig:student}
\end{figure}

Our student network (Figure~\ref{fig:student}) for single-image haze removal consists of cascaded blocks of residual connections, with channel-level and pixel-level attention to fuse semantics across all dimensions of the learned feature representation. The dehazing network is a deep, narrow network with a nested block structure to learn hierarchical representations at the feature level. It consists of \textit{inner} blocks with learnable convolutions, channel-level, and pixel-level attention, along with skip connections for residual learning. Furthermore, the \textit{outer} blocks interleave the inner blocks with extra learnable convolutions and skip connections to extract contextual information from different scales by mixing different receptive fields. The nested structure enables the capture of intra-stage multi-level features at a relatively reduced computational load. Multi-scale local and global features are extracted from feature maps of different resolutions.

\vspace{0.2em}

Each \textit{outer} block is constructed with $6$ \textit{inner} blocks to increase the depth and expressivity of the learned feature representation. Multiple convolutional layers are appended after each outer block with shortcut learning through skip connections to ensure information flow through the deep network. These connections are essential to retaining the low-density haze features as passed through. The \textit{inner} blocks with the residual connections, focusing on the locally significant regions of the feature map, also allow low-frequency haze regions to bypass further into the network, while the channel-level and pixel-level attention focus on the dense haze regions of the image.

\vspace{0.2em}

\change{Since the network is deep, skip (or) residual connections enable direct gradient flow for effective model training, avoiding the risks of exploding and vanishing gradients due to non-linear mappings. For an arbitrary input $\mathbf{X}$, the skip connections model $\mathcal{F}(\mathbf{X})+\mathbf{X}$, where $\mathcal{F}$ represents a function that is most commonly a combination of Convolution and ReLU (Rectified Linear Unit) activation layers. Following this, we implement channel-level and pixel-level attention to strengthen the network's representational capacity and effectively handle the channel and spatial information together.}

\vspace{0.2em}

\change{Given an input feature map $\mathbf{X}$ of $C$ channels, we perform the channel-level attention operation as \begin{equation}
 H^{c}=\operatorname{MLP} \left( \overbrace{\frac{1}{n \times m} \sum_{i} \sum_{j}}^{\text {GAP}} \mathbf{X^{c}}(i, j) \right)
\end{equation}}

\vspace{0.2em}

\change{where $\mathbf{X^{c}}(i, j)$ refers to the $(i, j)^{th}$ pixel location in the $m$ x $n$ feature map of channel $c$ in $\mathbf{X}$, GAP stands for Global Average Pooling~\cite{lin2013network}, and MLP refers to a Multi-Layer Perceptron. For implementation purposes, we use a two-layer MLP following the typical structure of Convolution, ReLU, and Sigmoid activation. Denoting $\odot$ as the Hadamard product (element-wise multiplication), the output $O^{c}$ obtained from the channel-level attention block can be represented as: \begin{equation}
O^{c}={H}^{c} \odot \mathbf{X}^{c} \
\end{equation}}

\vspace{0.2em}

\change{The output from the channel-level attention block then undergoes pixel-level attention, to further study spatial semantics. Pixel-level attention can simply be modelled using a two-layer MLP as used previously. The final output $P^{c}$ is:\begin{equation}
P^{c}=\operatorname{MLP}({O}^{c}) \odot {O}^{c} \
\end{equation}}

\vspace{0.2em}

Within each \textit{inner} block, the feature maps along the channel dimension are fused. Channel-level feature attention produces adaptive weights to learn the importance of channel-wise semantics. It works in coherence with the residual connections and pixel-level feature attention to bypass low-density haze (non-important semantics) while strongly emphasizing high-density, thick haze regions of the input (important semantics): essential to learn the texture required to remove the overlying haze. Through this mechanism, the network learns feature maps adaptively with different weights while focusing on densely hazed regions of the image, as most natural images occur with an uneven haze distribution.

% \subsection{Feature Affinity Module}
 
 \subsection{Rich Feature Distillation: Feature Affinity Module}

\begin{figure}[htbp]
\centering
\includegraphics[width=\textwidth]{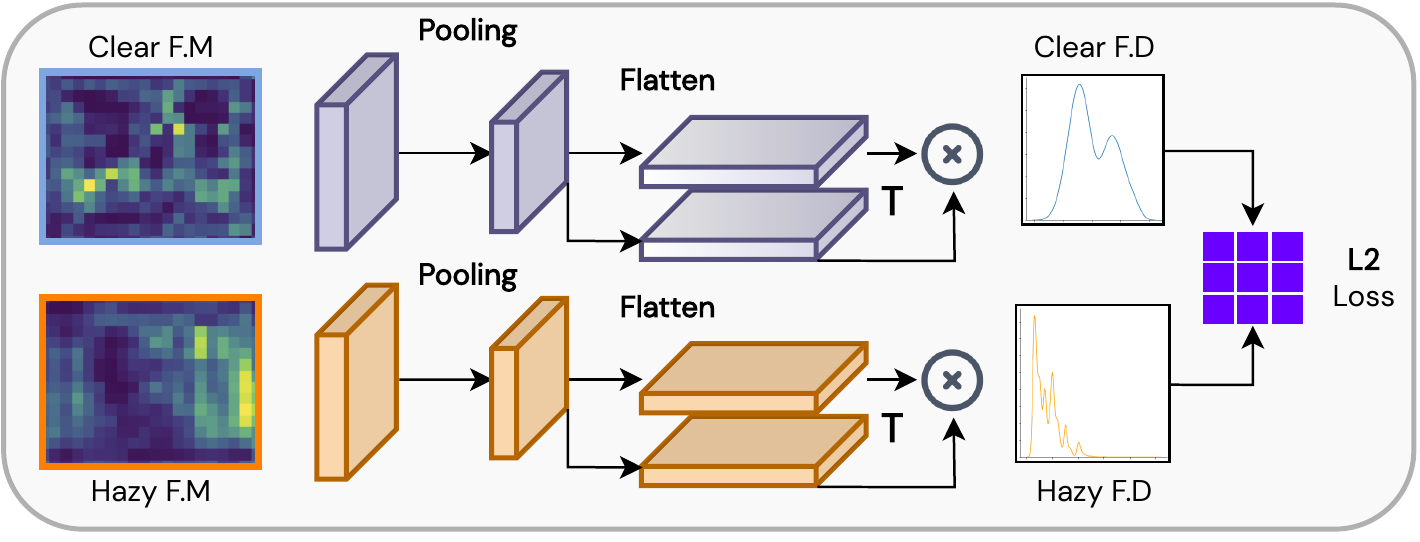}
\caption{\small Our feature affinity module to align the feature distribution and guide the student network. \emph{F.M} stands for \textbf{F}eature \textbf{M}ap. \emph{F.D} stands for \textbf{F}eature \textbf{D}istribution. \emph{$L_{2}$} refers to the Euclidean distance metric. \textbf{K}ullback-\textbf{L}eibler divergence can also be used, as a metric to measure distribution shift. \emph{T} refers to the \textbf{T}ranspose operation.}
\label{fig:FA_module}
\end{figure}

We designed an intelligent \textit{feature affinity module} at each scale to ensure that spatial and channel-aware features are distilled to make the best use of the higher resolution maps. Through this module shown in Figure~\ref{fig:FA_module}, the super-resolution teacher network features are used as “dark-knowledge supervision" to guide the process of representation learning for the student dehazing network. The clear and hazy intermediate feature maps from the teacher and student networks, \change{$C^{*}$ and $H^{*}$} respectively, are pooled by a scale of $0.25$ times (against the input map resolution). These pooled features are flattened spatially, with one resultant tensor per input channel. In order to account for spatial and channel-aware features in the distribution, the flattened spatial tensor and its transpose undergo a batch matrix product. \change{The sequence of operations is illustrated below:  \begin{multicols}{2}
  \begin{equation}
    {C^{*} = Flatten(AvgPool(C^{*}))}\ 
  \end{equation}\break
  \begin{equation}
    {H^{*} = Flatten(AvgPool(H^{*}))}\ 
  \end{equation}
\end{multicols}}

\vspace{-4em}
\change{\begin{multicols}{2}
  \begin{equation}
    {C = C^{*} \times {C^{*}}^{T}}\ 
  \end{equation}\break
  \begin{equation}
    {H = H^{*} \times {H^{*}}^{T}}\ 
  \end{equation}
\end{multicols}}

From the obtained distribution of spatial and channel-aware features, the Kullback-Leibler \cite{inbook} divergence is enforced to guide the hazy student model representation towards the clear teacher feature representation. \change{With the} clear distribution \emph{C} and hazy distribution \emph{H}, the KL divergence between the two can be estimated as:

\begin{equation}
  KL(H || C) = \sum_{}^{}H \log{\frac{H}{C}}  
\end{equation}

% \vspace{-0.5em}

Additionally, we also experiment with pixel-wise conformation between the intermediate feature representation and the $L_{2}$ loss function. The $L_{2}$ loss function builds on the Euclidean distance metric at the pixel level to calculate the dissimilarity between the clear and hazy image feature representations to guide the student network to effectively dehaze the images in real-time.

\begin{equation}
Pixel(H, C)=\frac{1}{n} \frac{1}{m} \sum_{i=1}^{n} \sum_{j=1}^{m}{(H(i, j)-C(i, j))}^{\change{2}}
\end{equation}

$C(i, j)$ corresponds to the $(i, j)^{th}$ pixel location in the $m$ x $n$ feature map of the clear image features. Similarly, $H(i, j)$ corresponds to the $(i, j)^{th}$ pixel location in the $m$ x $n$ feature map of the hazy image features. The feature affinity module loss function is denoted as $L_{FA}$ to ensure consistency.

\section{Experimental setup}

\subsection{Dataset Description}

% \vspace{1em}
\begin{figure}[htbp]
\centering
\includegraphics[width=0.9\textwidth]{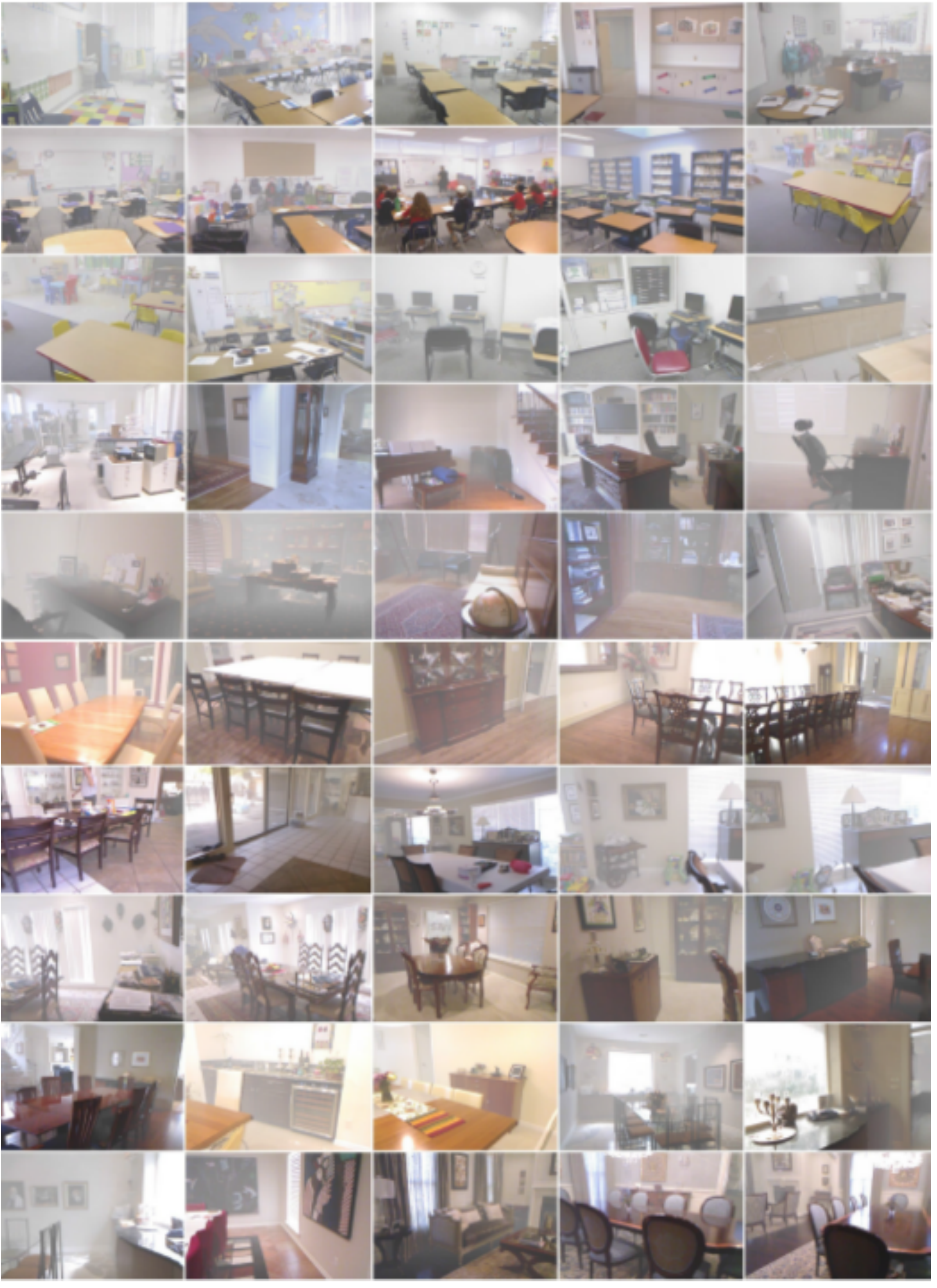}
\caption{The REalistic Single Image DEhazing (RESIDE) dataset. All our experiments utilize the RESIDE-Standard's Indoor Training Set (ITS) for training and RESIDE-Standard's Synthetic Objective Testing Set (SOTS) for testing the performance of our final network.}
\label{fig:dataset}
\end{figure}

For training our proposed framework and for validation required for hyperparameter tuning, the widely used RESIDE-Standard's Indoor Training Set (ITS) \cite{li2019benchmarking} is chosen. The REalistic Single Image DEhazing (RESIDE) dataset is a large-scale benchmark consisting of synthetic and real-world hazy images, as provided in Figure~\ref{fig:dataset}. The dataset entails diverse data sources and image content. RESIDE-Standard's Indoor Training Set (ITS) consists of 1399 clear images and corresponding 13990 hazy images with 13990 trans images. To test our trained model and verify the efficacy of our method, the RESIDE-Standard's Synthetic Objective Testing Set (SOTS) \cite{li2019benchmarking} is used. It consists of 500 indoor and outdoor subsets of clear and hazy images (50 clear and 500 hazy images each). The RESIDE dataset has been used across several works in the recent past \cite{DBLP:conf/aaai/QinWBXJ20, DBLP:conf/cvpr/HongXLQ20, d8d8e67ddd7547feaa86012179f83521}, and this work follows the same standard train-test split.

\subsection{Performance Metrics}

To implement our proposed framework, the super-resolution teacher network is first pre-trained. For this purpose, a suitable pixel-level loss function is used to train the network to converge during the optimization process. The Mean Squared Error (MSE) loss function minimizes the sum squared difference between the high-resolution ground truth and the super-resolution teacher network ($T$) output. For a given clear ground truth image $C_{HR}$ and a clear input low-resolution image $C_{LR}$, we obtain the error for a $H$ x $W$ image as follows:

\begin{equation}
L_{MSE (T)}=\frac{1}{H} \frac{1}{W} \sum_{i=1}^{H} \sum_{j=1}^{W}(C_{HR}(i, j)-T(C_{LR}(i, j)))^{2}
\end{equation}

Since we perform extensive experimentation on the standalone teacher and student networks, the student dehazing network is also trained independently. Hence, following contemporary works~\cite{DBLP:conf/aaai/QinWBXJ20, Cai2016DehazeNetAE} which have been established as baselines for single-image haze removal, the pixel-level optimization objective is adopted for the clear low-resolution ground truth $C_{LR}$ and the student dehazing network ($S$) output, given hazy input $H_{LR}$.

\begin{equation}
L_{MSE (S)}=\frac{1}{H} \frac{1}{W} \sum_{i=1}^{H} \sum_{j=1}^{W}(C_{LR}(i, j)-S(H_{LR}(i, j)))^{2}
\end{equation}

Our framework uses guiding features from the frozen super-resolution teacher network. Once the teacher network $T$ is frozen, the loss function adopted to carry out the final optimization of the combined network is a weighted combination of the MSE loss from the student dehazing network $L_{MSE (S)}$ and the feature affinity loss $L_{FA}$.

\begin{equation}
L_{Total} = L_{MSE (S)} + w_{FA} * L_{FA}
\end{equation}

where $w_{FA}$ is an adaptively chosen weight for the feature guidance from the teacher network.

To ensure consistency with prior works on single-image haze removal~\cite{Cai2016DehazeNetAE, DBLP:journals/tip/ZhangT20, ICCV17a}, two metrics are adopted: PSNR and Structural Similarity  Index (SSIM). The peak signal-to-noise ratio (PSNR) is a quantitative estimate of the ratio between the maximum possible power of an image and the power of corrupting noise that affects the quality of its representation. To estimate the PSNR of an image, it is necessary to compare that image to an ideal clean image with the maximum possible power. SSIM is used to measure the similarity between two given images. The SSIM is a perceptual metric that quantifies image quality degradation caused by processing such as data compression or losses in data transmission. It is a full reference metric that requires two images from the same image capture: a reference image and a processed image.

\subsection{Implementation Details}

For implementing our proposed framework for single-image haze removal, Python3 is used with PyTorch. All the experiments are carried out using a system with an Nvidia GeForce RTX 3080 Ti 24GB GPU and 64GB RAM. The models are trained using the Adam optimizer~\cite{kingma2017adam}. The super-resolution teacher model is pre-trained on the RESIDE training split for $20$ epochs. The teacher network is frozen to train the student dehazing network, from which rich features are extracted using our custom feature attention module. The student model is trained for $200$ epochs, with a batch size of 2. The learning rate at epoch = $0$ is set to $1e^{-5}$ and decays by 0.98 every 10 epochs. The adaptively chosen weight for the feature affinity loss, $w_{FA}$ is set to $0.25$. The hazy input images and corresponding clear images are maintained at their original resolution of $460$ x $620$. For a fair comparison, all benchmarked SOTA models (obtained from the author's official GitHub implementations) are re-implemented and re-trained on the same dataset.

\section{Results and Discussion}
\subsection{Teacher Network: Image Super-Resolution}

The performance of our proposed network for single-image super-resolution is compared to other baselines \cite{10.1007/978-3-319-46475-6_25, ledig2017photorealistic, guo2020closedloop} in Tables~\ref{tab:results_teacher1}, \ref{tab:results_teacher2}, \ref{tab:results_teacher3} and \ref{tab:results_teacher4}. Our experiments demonstrate the efficacy of our method for two HR image resolutions: \textsc{(460, 620)} in Tables ~\ref{tab:results_teacher1} and \ref{tab:results_teacher3}, \textsc{(920, 1240)} in Tables ~\ref{tab:results_teacher2} and \ref{tab:results_teacher4}. It can be inferred from Tables ~\ref{tab:results_teacher1} and \ref{tab:results_teacher2} that our method, \textbf{MC-FSRCNN} has the best compute-performance trade-off based on the number of parameters vs. PSNR. As we move to the HR image resolution of \textsc{(920, 1240)}, our model outperforms both baselines.

\begin{table}[!h]
\renewcommand{\arraystretch}{2}
\centering
\caption{Comparison of lightweight single-image super-resolution networks, against our proposed teacher network [HR Image Resolution: \textbf{(460, 620)}]}
\begin{tabular}{|c|c|c|c|}
%\begin{tabular}{lcccp{2.5cm}p{2.5cm}p{2.5cm}p{2.5cm}}
\hline
\textbf{Model} & \textbf{Num. Parameters} & \textbf{PSNR} & \textbf{Epoch} \\
\hline
FSRCNN-2x~\cite{10.1007/978-3-319-46475-6_25} & 24.68K & 40.22 & 19 \\
\hline
FSRCNN-4x~\cite{10.1007/978-3-319-46475-6_25} & 24.68K & 29.31 & 19 \\
\hline
SR-ResNet~\cite{ledig2017photorealistic} & 1.5M & 30.45 & 15 \\ \hline
DRN-S~\cite{guo2020closedloop} & 4.8M & 58.21 & 13 \\ \hline
\textbf{MC-FSRCNN} & 38.42K & 40.53 & 12 \\
\hline
\end{tabular}
\label{tab:results_teacher1}
\end{table}

\begin{table}[!h]
\renewcommand{\arraystretch}{2}
\centering
\caption{Comparison of lightweight single-image super-resolution networks, against our proposed teacher network [HR Image Resolution: \textbf{(920, 1240)}]}
\begin{tabular}{|c|c|c|c|}
%\begin{tabular}{lcccp{2.5cm}p{2.5cm}p{2.5cm}p{2.5cm}}
\hline
\textbf{Model} & \textbf{Num. Parameters} & \textbf{PSNR} & \textbf{Epoch} \\
\hline
FSRCNN-2x~\cite{10.1007/978-3-319-46475-6_25} & 24.68K & 60.02 & 18 \\
\hline
DRN-S~\cite{guo2020closedloop} & 4.8M & 59.27 & 14 \\
\hline
\textbf{MC-FSRCNN} & 38.42K & \textbf{67.18} & 13 \\
\hline

\end{tabular}
\label{tab:results_teacher2}
\end{table}

Furthermore, to study the choice of network modules, a minimal ablation study for each HR image resolution is conducted in Table~\ref{tab:results_teacher3} and \ref{tab:results_teacher4}. It is observed that the network is very sensitive to the choice of activation function \cite{agarap2019deep} (default: Parametric ReLU) and to the placement of batch normalization \cite{ioffe2015batch}. Finally, \textit{Depthwise Separable Convolutions} are leveraged to reduce the model size even further. The resulting model (\textbf{DSC-MC-FSRCNN}) exhibits the most optimal compute-performance trade-off, with a reduction of 31.15\% in model size.  

\begin{table}[!h]
\renewcommand{\arraystretch}{2}
\centering
\caption{HR Image Resolution: \textbf{(460, 620)}, \textbf{MC-FSRCNN} Ablation Variants}
\begin{tabular}{|c|c|c|c|}
%\begin{tabular}{lcccp{2.5cm}p{2.5cm}p{2.5cm}p{2.5cm}}
\hline
\textbf{Model} & \textbf{Num. Parameters} & \textbf{PSNR} & \textbf{Epoch} \\
\hline
ReLU-MC-FSRCNN & 38.42K & 40.15 & 13 \\
\hline
DSCBN-MC-FSRCNN\footnotemark & 29.57K & 38.12 & 16 \\
\hline
\textbf{DSC-MC-FSRCNN} & 26.45K & \textbf{40.12} & 19 \\
\hline
\end{tabular}
\label{tab:results_teacher3}
\end{table}

\begin{table}[!h]
\renewcommand{\arraystretch}{2}
\centering
\caption{HR Image Resolution: \textbf{(920, 1240)}, \textbf{MC-FSRCNN} Ablation Variants}
\begin{tabular}{|c|c|c|c|}
%\begin{tabular}{lcccp{2.5cm}p{2.5cm}p{2.5cm}p{2.5cm}}
\hline
\textbf{Model} & \textbf{Num. Parameters} & \textbf{PSNR} & \textbf{Epoch} \\
\hline
ReLU-MC-FSRCNN & 38.42K & 62.96 & 17 \\
\hline
DSCBN-MC-FSRCNN\footnotemark & 29.57K & 43.07 & 17 \\
\hline
\textbf{DSC-MC-FSRCNN} & 26.45K & \textbf{65.94} & 16 \\
\hline
\end{tabular}
\label{tab:results_teacher4}
\end{table}

\footnotetext[1]{DSC refers to Depthwise Separable Convolution~\cite{8099678}}

\begin{figure}[htbp]
  \begin{center}
  \subfloat[Output]{\label{fig:mcf_o}\includegraphics[width=0.375\textwidth]{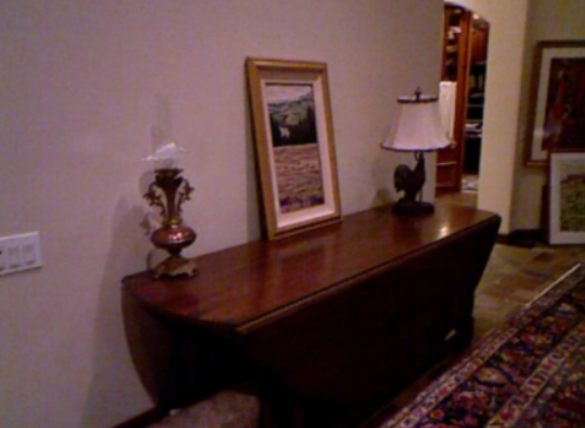} }
      ~ 
    \subfloat[High-Resolution GT]{\label{fig:mcf_hrgt}\includegraphics[width=0.375\textwidth]{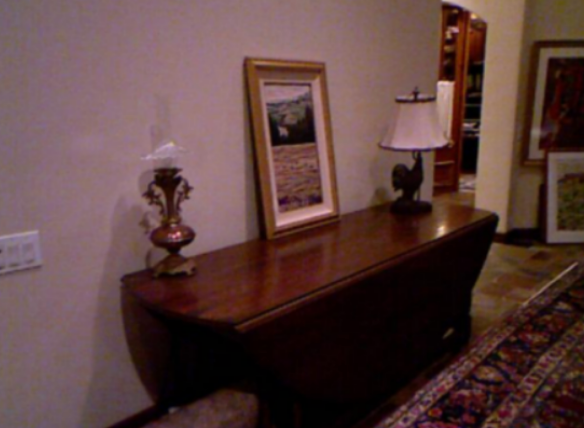}}

    \caption{MC-FSRCNN, \textbf{PSNR:} 67.18 dB}
    \label{fig:mcf}
        \end{center}
  \end{figure}

\begin{figure}[htbp]
  \begin{center}
  \subfloat[Output]{\label{fig:dsc_mcf_o}\includegraphics[width=0.375\textwidth]{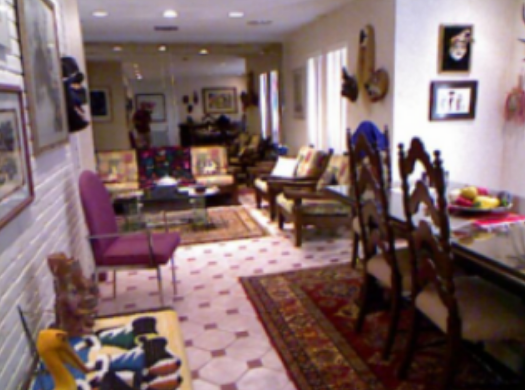} }
      ~ 
    \subfloat[High-Resolution GT]{\label{fig:dsc_mcf_hrgt}\includegraphics[width=0.375\textwidth]{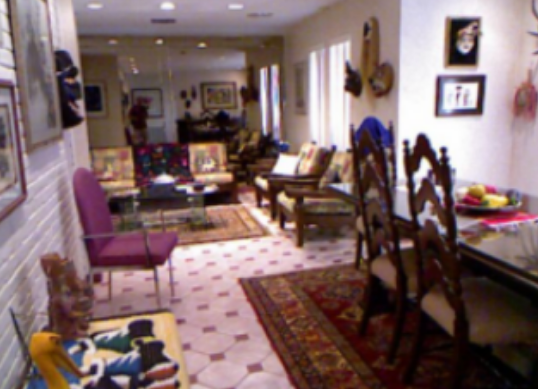}}

    \caption{DSC-MC-FSRCNN, \textbf{PSNR:} 65.94 dB}
    \label{fig:dsc_mcf}
        \end{center}
  \end{figure}

Figure~\ref{fig:mcf} and ~\ref{fig:dsc_mcf} provide a visual representation of the obtained results using \textbf{MC-FSRCNN} and \textbf{DSC-MC-FSRCNN} respectively. The overlay in the above figures, to compare each model against their high-resolution GT (Ground-Truth) images, confirms that the proposed approach is indeed visually robust to various scenes. The competitive PSNR values are obtained with just a few tens of thousands of parameters, showing great promise for lightweight haze-removal frameworks that can be efficiently deployed on the edge while exploiting these rich super-resolution teacher features.

% \begin{figure}[htbp]
%   \begin{center}
%   \subfloat[Output]{\label{fig:mcf_o}\includegraphics[width=0.3\textwidth]{images/m2_o1.png} }
%   \vfil
%     \subfloat[High-Resolution GT]{\label{fig:mcf_hrgt}\includegraphics[width=0.3\textwidth]{images/m2_GT.png}}

%     \caption{MC-FSRCNN, \textbf{PSNR:} 67.18 dB}
%     \label{fig:mcf}
%         \end{center}
%   \end{figure}

% \begin{figure}[htbp]
%   \begin{center}
%   \subfloat[Output]{\label{fig:dsc_mcf_o}\includegraphics[width=0.475\textwidth]{images/m1_o1.png} }
%     \vfil
%     \subfloat[High-Resolution GT]{\label{fig:dsc_mcf_hrgt}\includegraphics[width=0.475\textwidth]{images/m1_GT.png}}

%     \caption{DSC-MC-FSRCNN, \textbf{PSNR:} 65.94 dB}
%     \label{fig:dsc_mcf}
%         \end{center}
%   \end{figure}

% \vspace{-1em}

\subsection{Student Network: Single Image Haze Removal}

  \begin{figure}[!h]
  \centering
  \includegraphics[width=10 cm,height=11 cm] {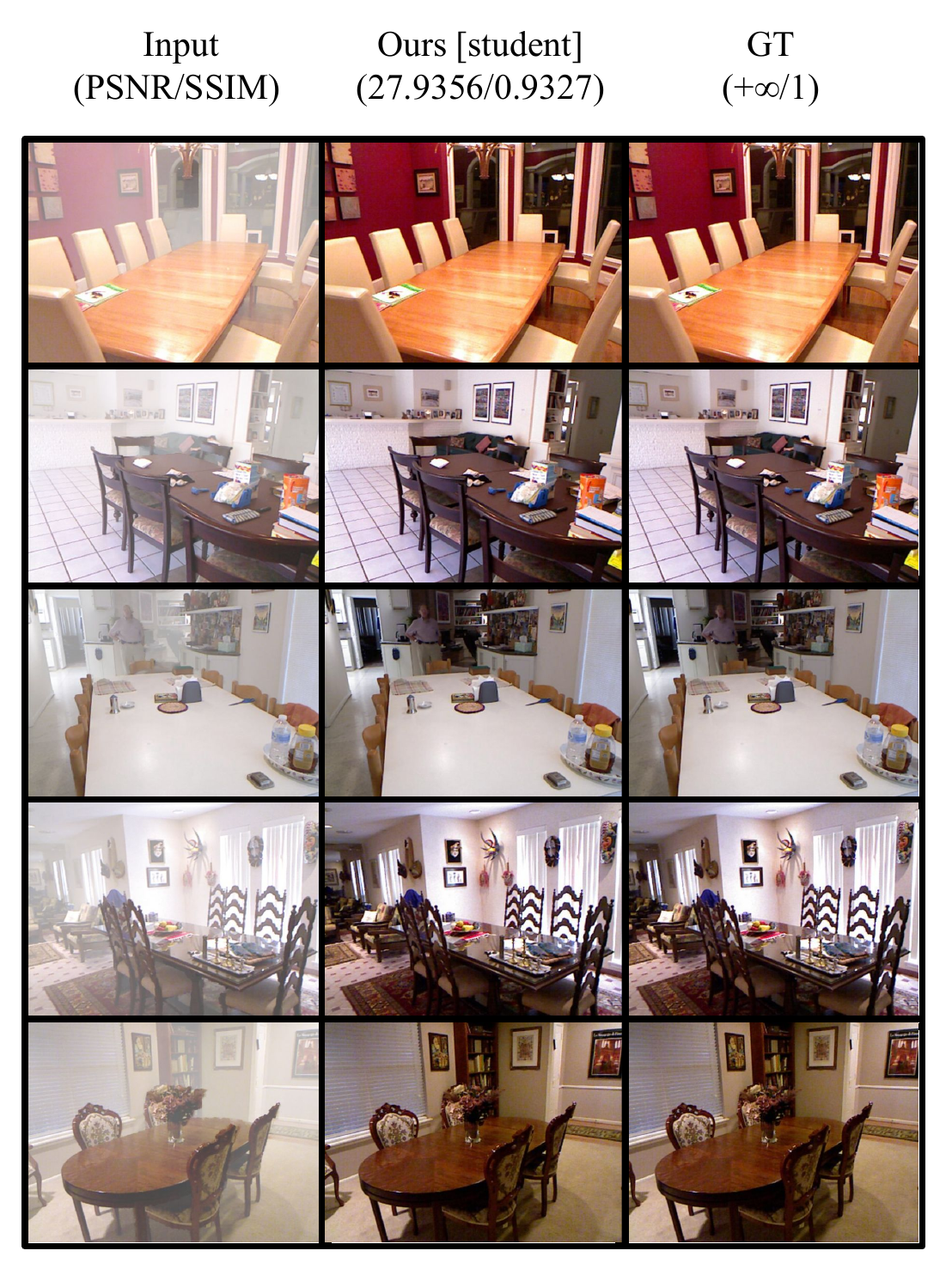}
\caption{\change{Qualitative comparison to evaluate student model performance on RESIDE SOTS (indoor) (test set). Row-wise, \textbf{Top:} hazy input images, \textbf{Middle:} clear output images of our student network (dehazed), \textbf{Bottom:} clear ground-truth images.}}
\label{fig:student_qualitative}
\end{figure}

For single image haze removal baselines, several established previous works are reported in Table~\ref{tab:main_results}. It is inferred that our method has the best compute-performance trade-off based on the number of parameters vs. PSNR and SSIM. Table~\ref{tab:main_results} reports the performance of existing works upon training on RESIDE-ITS and testing on RESIDE-SOTS (Indoor) \cite{li2019benchmarking}. The various frameworks included in the table are based on a fixed compute budget. For instance, GCANet \cite{chen2018gated} obtains a marginal improvement over our proposed method in terms of PSNR and SSIM, but comes with an increase in computing cost since the GCANet model is \emph{7x} bigger than ours. Given the importance of lightweight models to be deployed on the edge, our method is memory and compute efficient while also exhibiting improved performance than other lightweight counterparts.

% \vspace{-1.5em}
% \begin{table}[htbp]
% \renewcommand{\arraystretch}{2}
% \centering
% \caption{Comparison of lightweight single-image haze removal networks, against our proposed student network}
% \begin{tabular}{|c|c|c|c|}
% %\begin{tabular}{lcccp{2.5cm}p{2.5cm}p{2.5cm}p{2.5cm}}
% \toprule
% \textbf{Model} & \textbf{\# Params (Mil.)}   & \textbf{PSNR}    & \textbf{SSIM}   \\ \hline
% % FPCNet         & 288                  & 21.841            & 0.887          \\ \hline
% AOD-Net \cite{ICCV17a}       & 0.00176                & 19.064            & 0.850          \\ \hline
% LSID-FCNN \cite{article}      & 0.00429                & 19.530          & 0.836          \\ \hline
% MSCNN \cite{Ren2016SingleID}         & 0.00801                & 17.572            & 0.810          \\ \hline
% DeHazeNet \cite{Cai2016DehazeNetAE}     & 0.00831                & 19.826            & 0.821          \\ \hline
% FAMED-Net \cite{DBLP:journals/tip/ZhangT20}     & 0.018                  & 27.014            & 0.937          \\ \hline
% GFN \cite{Ren2018GatedFN}           & 0.514                & 24.918            & 0.919          \\ \hline
% GCANet \cite{chen2018gated}       & 0.703                & 30.230            & 0.980            \\ \hline
% DCPDN \cite{Zhang_2018_CVPR}         & 66.9                & 20.811            & 0.838          \\ \hline
% \textbf{Ours (Student)}  & \textbf{0.096} & \textbf{27.9356} & \textbf{0.9327} \\
% \bottomrule
% \end{tabular}
% \label{tab:main_results}
% \end{table}

\begin{table}[htbp]
\renewcommand{\arraystretch}{1.54}
\centering
\caption{\change{Comparison of single-image haze removal methods, against our proposed framework}}
\begin{tabular}{|c|c|c|c|}
%\begin{tabular}{lcccp{2.5cm}p{2.5cm}p{2.5cm}p{2.5cm}}
\toprule
\textbf{Model} & \textbf{\# Params (Mil.) ($\downarrow$)}   & \textbf{PSNR ($\uparrow$)}    & \textbf{SSIM ($\uparrow$)}   \\ \hline
% FPCNet         & 288                  & 21.841            & 0.887          \\ \hline
AOD-Net \cite{ICCV17a}       & 0.00176                & 19.064            & 0.850          \\ \hline
SSID \cite{8902220}       & 9.232               & 20.959            & 0.814          \\ \hline
ZID \cite{9170880}       & 42.951               & 19.830            & 0.835          \\ \hline
LSID-FCNN \cite{article}      & 0.00429                & 19.530          & 0.836          \\ \hline
MSCNN \cite{Ren2016SingleID}         & 0.00801                & 17.572            & 0.810          \\ \hline
DeHazeNet \cite{Cai2016DehazeNetAE}     & 0.00831                & 19.826            & 0.821          \\ \hline
FAMED-Net \cite{DBLP:journals/tip/ZhangT20}     & 0.018                  & 27.014            & 0.937          \\ \hline
GFN \cite{Ren2018GatedFN}           & 0.514                & 24.918            & 0.919          \\ \hline
GCANet \cite{chen2018gated}       & 0.703                & 30.230            & 0.980            \\ \hline
DCPDN \cite{Zhang_2018_CVPR}         & 66.902                & 20.811            & 0.838          \\ \hline
Cycle-Dehaze \cite{engin2018cycle}         & 10.602               & 18.880            & 0.810          \\ \hline
MSBDN \cite{https://doi.org/10.48550/arxiv.2004.13388}         & 31.353               & 28.341            & 0.955          \\ \hline
FFA-Net \cite{DBLP:conf/aaai/QinWBXJ20}         & 2.871                & 32.178            & 0.978          \\ \hline
PFDN \cite{10.1007/978-3-030-58577-8_12}         & 11.278                & 32.68            & 0.976          \\ \hline
4K-Dehazing \cite{9578433}     & 34.547               & 23.881            & 0.932          \\ \hline
GridDehazeNet \cite{liuICCV2019GridDehazeNet}         & 0.956                & 28.341            & 0.962          \\ \hline
Ours [Student] & 0.096 & 27.9356 & 0.9327 \\
\hline
\textbf{Ours} & 0.122 & 32.1412 & 0.9666 \\
\bottomrule
\end{tabular}
\label{tab:main_results}
\end{table}

In order to provide a qualitative, visual depiction of our proposed method's performance, the dehazed model output is compared to the clear ground truth images in Figure~\ref{fig:student_qualitative}. It provides five hazy sample images in the first row from the RESIDE-SOTS indoor testing set as a reference benchmark for measuring model performance. Furthermore, the second row consists of the outputs of our student model for the corresponding hazy images above. To emphasize the highly performant nature of our lightweight, efficient network, the corresponding clear ground truth images of the hazy counterparts are provided in the third row. Given that the second and third rows are highly similar from a visual and perceptual perspective, it is inferred that our proposed method can be leveraged for various on-the-edge downstream tasks that require clear images from hazy environments.

\subsection{Complete Framework: Single Image Haze Removal}

In this subsection, the results obtained for our complete framework are provided, compared to other baselines and our proposed standalone student dehazing network. Here, the super-resolution teacher network is frozen after pre-training. Using the obtained clear image feature representation, the outputs obtained at multiple scales are forwarded to the feature affinity module in coherence with the student network's intermediate hazy feature distribution. The student network and the components of the feature affinity module are trained in the final stage of optimization until the convergence of the student dehazing network.

 \begin{figure}[!h]
 \centering
\includegraphics[width=10 cm,height=11 cm]{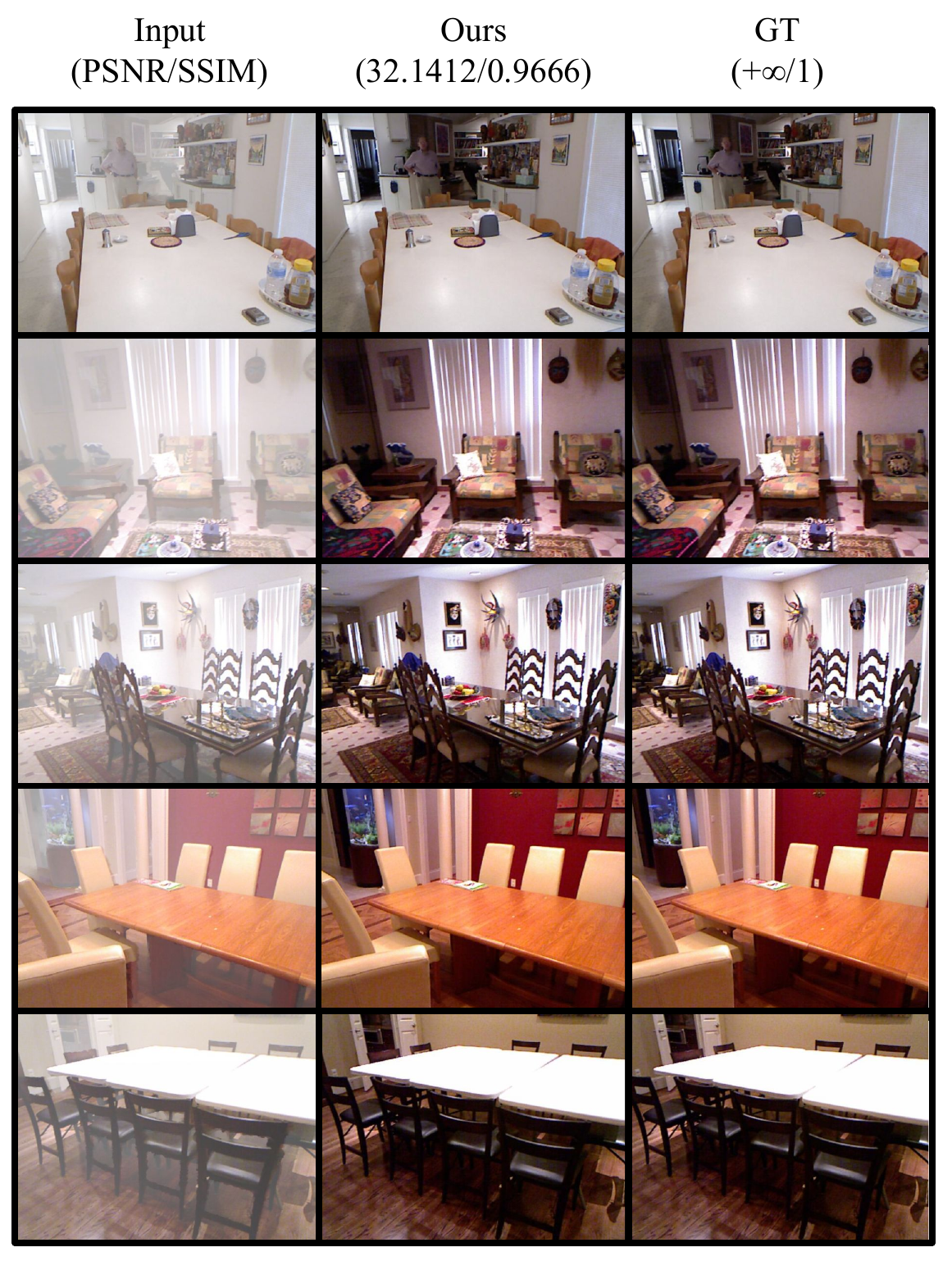}
\caption{\change{Qualitative comparison to evaluate the performance of our framework on RESIDE SOTS (indoor) (test set). Row-wise, \textbf{Top:} hazy input images, \textbf{Middle:} clear output images of our framework (dehazed), \textbf{Bottom:} clear ground-truth images.}}
\label{fig:full_qualitative}
\end{figure}

The qualitative results of our framework are depicted in Figure~\ref{fig:full_qualitative}. The quantitative results are presented in Table~\ref{tab:main_results}, to verify the improved performance of our framework over the baselines. It is noteworthy that at the cost of adding $26K$ new parameters, there is a significant performance gain for single image haze removal in terms of SSIM and PSNR.

% % \hspace{-3em}
% \begin{table}[htbp]
% \renewcommand{\arraystretch}{2}
% \centering
% \caption{Results for the complete framework - Efficient Single Image Haze Removal}
% \begin{tabular}{|c|c|c|c|}
% %\begin{tabular}{lcccp{2.5cm}p{2.5cm}p{2.5cm}p{2.5cm}}
% \hline
% \textbf{Model} & \textbf{Num. Parameters (Mil.)} & \textbf{PSNR} & \textbf{SSIM} \\
% \hline
% FAMED-Net~\cite{DBLP:journals/tip/ZhangT20} & 0.018     & 27.014       & 0.937   \\
% \hline
% GCANet \cite{chen2018gated}       & 0.703                & 30.230            & 0.980            \\ 
% \hline
% DCPDN \cite{Zhang_2018_CVPR}         & 66.9                & 20.811            & 0.838          \\ 
% \hline
% Student & 0.096 & 27.9356 & 0.9327 \\
% \hline
% \textbf{Ours (Full)} & 0.122 & 32.1412 & 0.9666 \\
% \hline
% \end{tabular}
% \label{tab:main_results}
% \end{table}

\subsection{Performance analysis of the model}

\subsubsection{Effect of teacher feature guidance resolution}
The teacher network for super-resolution plays an essential role in feature distillation, enabling the lightweight student dehazing network to learn a semantically accurate clear feature distribution, given hazy inputs. We demonstrate the effect of using lower resolution teacher features ($460$ x $620$) and using higher resolution teacher features ($920$ x $1240$) on the dark knowledge supervision being passed on to the student dehazing network.

% \vspace{1em}

From Table~\ref{tab:ablation_1} it is inferred that the higher resolution features provide a marginal improvement in PSNR and a substantial improvement in SSIM, to guide the clear feature representation learning process.

\begin{table}[htbp]
\renewcommand{\arraystretch}{2}
\centering
\caption{Effect of Teacher Feature Resolution. Loss function-$L_{2}$ and $w_{FA}$=0.25, based on final model}
\begin{tabular}{|c|c|c|}
%\begin{tabular}{|p{2.5cm}|p{2.5cm}|p{2cm}|}
\hline
\textbf{\change{Resolution}} & \textbf{PSNR} & \textbf{SSIM} \\
\hline
Low-Resolution (LR) & 31.9616 & 0.9508 \\
\hline
\textbf{High-Resolution (HR)} & 32.1412 & 0.9666 \\
\hline
\end{tabular}
\label{tab:ablation_1}
\end{table}

\subsubsection{Effect of $w_{FA}$ from the feature affinity module}

For the overall network optimization, the loss function is a weighted combination of the MSE loss from the student dehazing network $L_{MSE (S)}$ and the feature affinity loss $L_{FA}$. 

\begin{equation}
L_{Total} = L_{MSE (S)} + w_{FA} * L_{FA}
\end{equation}

\begin{table}[!htb]
\renewcommand{\arraystretch}{2}
\centering
\caption{Effect of $w_{FA}$ value. Loss function-$L_{2}$ and resolution set to HR, based on final model}
\begin{tabular}{|c|c|c|}
%\begin{tabular}{lcccp{2.5cm}p{2.5cm}p{2.5cm}p{2.5cm}}
%\begin{tabular}{|p{2.5cm}|p{2.5cm}|p{2cm}|}
\hline
\textbf{\change{$w_{FA}$}} & \textbf{PSNR} & \textbf{SSIM} \\
\hline
\textbf{0.25} & 32.1412 & 0.9666 \\
\hline
0.50 & 31.9939 & 0.9551 \\
\hline
0.75 & 32.2175 & 0.9603 \\
\hline
\end{tabular}
\label{tab:ablation_2}
\end{table}

In the above equation, $w_{FA}$ is an adaptively chosen weight for the feature guidance from the teacher network that requires tuning. Due to limited compute budget, experiments are carried out with three preset values for $w_{FA}$: 0.25, 0.5, and 0.75 respectively. From Table~\ref{tab:ablation_2}, it is evident that the value of $w_{FA}$ plays only a minor, insignificant role in the convergence of the network to optimal values of PSNR and SSIM. Based on the obtained results, the optimal value for the hyperparameter is set as 0.25.

\subsubsection{Effect of the choice of loss function for feature distillation}

To align the hazy feature representation distribution closer to the clear feature distribution, a loss function is adopted in the feature affinity module, referred to as the feature affinity loss $L_{FA}$. Three different loss functions are experimented with for $L_{FA}$, namely the KL-Divergence loss, $L_{1}$ loss, and the $L_{2}$ loss function.

\begin{table}[htbp]
\renewcommand{\arraystretch}{2}
\centering
\caption{Effect of Loss Function. $w_{FA}$=0.25 and resolution set to HR, based on final model}
\begin{tabular}{|c|c|c|}
%\begin{tabular}{lcccp{2.5cm}p{2.5cm}p{2.5cm}p{2.5cm}}
%\begin{tabular}{|p{2.5cm}|p{2.5cm}|p{2cm}|}
\hline
\textbf{\change{Loss Function}} & \textbf{PSNR} & \textbf{SSIM} \\
\hline
$L_{1}$ & 30.4805 & 0.9376  \\
\hline
KL Divergence & 30.1275 & 0.9187 \\
\hline
\textbf{$L_{2}$} & 32.1412 & 0.9666 \\
\hline
\end{tabular}
\label{tab:ablation_3}
\end{table}

From Table~\ref{tab:ablation_3}, it can be concluded that the choice of the chosen loss function $L_{FA}$ plays a considerable role in the obtained result. The obtained PSNR and SSIM values confirm a performance gain upon the usage of the $L_{2}$ loss function, based on the Euclidean distance metric.
\begin{table}[!h]
\centering
\renewcommand{\arraystretch}{2}
\caption{Computational Performance Metrics}
\begin{tabular}{|c|c|c|c|c|}
%\begin{tabular}{lcccp{2.5cm}p{2.5cm}p{2.5cm}p{2.5cm}}
\hline
\textbf{Model}                    & \textbf{Parameters } & \textbf{Size } & \textbf{GFLOPS} & \textbf{Inference} \\
                       &    (Mil.)           &  (MB)          &           & 
\textbf{Time} (ms) \\ \hline

AOD-Net \cite{ICCV17a}                        & 0.00176              & 0.011              & 0.998           & 6.4569                                  \\ \hline
FFA-Net \cite{DBLP:conf/aaai/QinWBXJ20}                        & 2.87               & 11.184             & 1613.042        & 249.9094                                \\ \hline
DeHazeNet \cite{Cai2016DehazeNetAE}                        & 0.00831               & 0.036              & 4.588           & 11.3885                                 \\ \hline
Grid-DeHazeNet  \cite{liuICCV2019GridDehazeNet}                  & 0.956               & 3.759              & 155.043         & 74.7072                                 \\ \hline
GCANet \cite{chen2018gated}                     & 0.703               & 2.713              & 126.301         & 59.6612                                 \\ \hline
MSBDN-DFF \cite{https://doi.org/10.48550/arxiv.2004.13388} & 31.353             & 119.786            & 165.968         & 111.5572                                \\ \hline
4K-Dehazing \cite{9578433}                     & 34.547             & 131.987            & 210.715         & 37.497                                  \\ \hline
Cycle-Dehaze \cite{engin2018cycle}         & 10.602             & 40.464             & 311.746         & 68.7084                                 \\ \hline
% KDDN   \cite{9156318}      & 22.99              & 87.815             & 112.932         & 19.9289                                 \\ \hline
% AECR-Net                          & 2.62M               & 9.994              & \textbf{TBD}    & \textbf{TBD}                            \\ \hline
\textbf{Ours}                     & 0.122        & 0.634       & 64.039  &  58.981                            \\ \hline
\end{tabular}%

\label{tab:compute}
\end{table}
\subsubsection{Computational Bench-marking}

Supporting our claim of computational efficiency, our work is compared to other single image haze removal frameworks that have produced significant advancements and utilization of cross-domain techniques. To record these computational benchmarks, we adopt the number of model parameters, model size (MB), GFLOPS, and end-to-end inference time as comparison metrics, in Table~\ref{tab:compute}.

\section{Conclusion}
This work proposed an efficient, lightweight pipeline for single image haze removal. The framework introduced exploits a rich pre-trained super-resolution teacher network to distill multi-scale information to the student dehazing network. The fine-grained features at different scales, as guided by the super-resolution network, are merged into the feature distribution of the student in an intelligent manner using a feature affinity module. Moving beyond prior works with high architectural and computational complexity, we utilized the notion of heterogeneous knowledge distillation, further highlighting the significance of rich teacher-feature guidance in solving the long-standing problem of haze removal. \change{As a result of this guidance, our lightweight, efficient student dehazing network achieves gains of upto 15\% in PSNR, and 4\% in terms of SSIM, while reducing the model size by $\sim$20 times. The considerable reduction in model size, complementing the performance gain arising from our feature affinity module and rich teacher features, sets a precedent for on-the-edge applications of single image haze removal. Given that our work ensures computational efficiency, cheap inference cost, and good performance, we hope to accelerate future research in improving the numerous applications of haze removal, such as remote sensing, intelligent vehicle control, and underwater image dehazing, amongst others.}

Our current work used a fully-supervised paradigm to train the network required for dark knowledge feature guidance. There is scope to explore cross-domain specialization techniques in the domain of self-supervised learning, possibly in a contrastive setting for single image haze removal. Furthermore, there is also scope for advancing related work in the unsupervised paradigm for haze removal. Given our attempt to align the hazy feature distribution towards the clear feature distribution in the multi-level intermediate representations, there is a possibility to proceed in the direction of adversarial guidance to bridge the distribution shift. Finally, there is a gap in the data distribution shift amongst models trained on synthetic, curated hazy data, but tested on real-world haze images. Real-world data domain adaptation still needs considerable improvement and is a direction to be explored.

\section*{Declaration of Competing Interest}
The authors declare that they have no conflict of interests.

% \section*{References}
\bibliographystyle{unsrt}  
{\footnotesize
\bibliography{dehaze}}

\end{document}